\documentclass[runningheads]{llncs}

\usepackage{eccv}

\usepackage{eccvabbrv}
\usepackage{multirow}
\usepackage{enumitem}
\usepackage{graphicx}
\usepackage{booktabs}
\usepackage{wrapfig}
\usepackage{siunitx}
\usepackage{xcolor}
\usepackage{tcolorbox}
\definecolor{MainPurple}{RGB}{102, 0, 153}
\definecolor{CodeBg}{RGB}{248, 248, 248}
\usepackage[accsupp]{axessibility}

\usepackage{hyperref}
\usepackage{orcidlink}

\begin{document}
    \title{Render-in-the-Loop: Vector Graphics Generation via Visual Self-Feedback}

    \author{Guotao Liang\inst{1} \and Zhangcheng Wang\inst{3} \and Juncheng Hu\inst{1} \and Haitao Zhou\inst{1} \and Ziteng Xue\inst{1} \and Jing Zhang\inst{1} \and Dong Xu\inst{2} \and Qian Yu\inst{1}\thanks{Corresponding author.}
    }

\authorrunning{G.~Liang et al.}

\institute{
School of Software, Beihang University, China \\
\email{\{liangguotao, hujuncheng, zhouhaitao, zt\_xue, zhang\_jing, qianyu\}@buaa.edu.cn}  \and
Department of Computer Science, The University of Hong Kong, China \\
\email{dongxu@cs.hku.hk}
\and 4Paradigm, China \\
\email{wzc1@mail.ustc.edu.cn}
}
\titlerunning{Render-in-the-Loop}
    \maketitle

    \begin{figure}[!h]
    \centering
    \includegraphics[width=\textwidth]{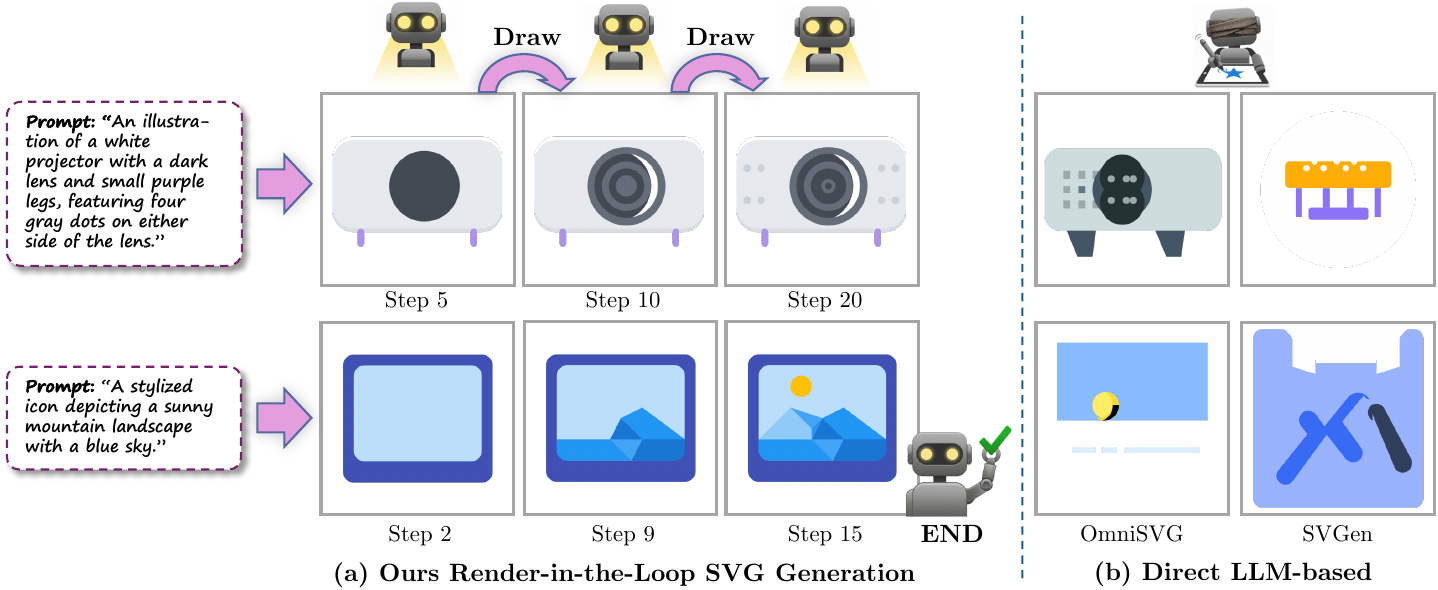}
    \caption{\textbf{Render-in-the-Loop Generation.} (a) Our \textbf{Visual Self-Feedback} explicitly renders intermediate code into a canvas, feeding it back to provide continuous visual guidance. (b) Traditional open-loop approaches draw ``blindly'' using only textual history, often struggling with geometric accuracy and visual quality. By closing the loop, our method ensures structurally coherent and high-quality SVG synthesis.}
    \label{fig:teaser}
    \end{figure}

    \begin{abstract}
Multimodal Large Language Models (MLLMs) have shown promising capabilities in generating Scalable Vector Graphics (SVG) via direct code synthesis. However, existing paradigms typically adopt an open-loop ``blind drawing'' approach, where models generate symbolic code sequences without perceiving intermediate visual outcomes. This methodology severely underutilizes the powerful visual priors embedded in MLLMs' vision encoders, treating SVG generation as a disjointed textual sequence modeling task rather than an integrated visuo-spatial one. Consequently, models struggle to reason about partial canvas states and implicit occlusion relationships, which are visually explicit but textually ambiguous. To bridge this gap, we propose \textbf{Render-in-the-Loop}, a novel generation paradigm that reformulates SVG synthesis as a step-wise, visual-context-aware process. By rendering intermediate code states into a cumulative canvas, the model explicitly observes the evolving visual context at each step, leveraging on-the-fly feedback to guide subsequent generation. However, we demonstrate that applying this visual loop naively to off-the-shelf models is suboptimal due to their inability to leverage incremental visual-code mappings. To address this, we first utilize fine-grained path decomposition to construct dense multi-step visual trajectories, and then introduce a Visual Self-Feedback (VSF) training strategy to condition the next primitive generation on intermediate visual states. Furthermore, a Render-and-Verify (RaV) inference mechanism is proposed to effectively filter degenerate and redundant primitives. Our framework, instantiated on a multimodal foundation model, outperforms strong open-weight baselines on the standard MMSVGBench. This result highlights the remarkable data efficiency and generalization capability of our Render-in-the-Loop paradigm for both Text-to-SVG and Image-to-SVG tasks. 
\keywords{Scalable Vector Graphics \and Multimodal Large Language Models \and Visual Self-Feedback}
\end{abstract}

    \section{Introduction}

Scalable Vector Graphics (SVG)~\cite{svg_w3c_1999}, characterized by resolution independence, high editability, and compact storage, have become an indispensable graphic format in modern UI/UX design, industrial typography, and front-end development~\cite{visualizing_geroimenko_2005, integrating_yan_2006, deepvecfont_wang_2021, wordasimage_iluz_2023}. With the rapid progress of Multimodal Large Language Models (MLLMs) in code generation and visual understanding~\cite{gpt4_achiam_2023,qwen3vl_bai_2025}, directly leveraging foundation models for \emph{Text-to-SVG} and \emph{Image-to-SVG} generation has recently emerged as a promising research direction~\cite{iconshop_wu_2023,starvector_rodriguez_2025,omnisvg_yang_2025}.

Pioneering works such as StarVector~\cite{starvector_rodriguez_2025} and OmniSVG~\cite{omnisvg_yang_2025} have demonstrated that MLLMs are capable of directly producing SVG XML code. However, existing generation paradigms typically adopt an open-loop ``blind drawing'' approach~\cite{iconshop_wu_2023,starvector_rodriguez_2025,omnisvg_yang_2025,llm4svg_xing_2025,svgthinker_chen_2025}, where the model generates symbolic code sequences without perceiving the intermediate visual outcomes. This methodology severely underutilizes the powerful visual priors embedded in MLLMs' vision encoders, treating SVG generation as a disjointed textual sequence modeling task rather than an integrated visuo-spatial one~\cite{llm4svg_xing_2025,chat2svg_wu_2025,svgen_wang_2025}. As noted in recent studies~\cite{rlrf_rodriguez_2025}, this disconnect between abstract code and its physical rendering often leads to overfitting on specific coordinate distributions, causing models to struggle with generalizing to diverse or out-of-distribution inputs. Without observing the canvas, models may hallucinate geometries that are syntactically valid but visually incoherent, and struggle to reason about partial canvas states and implicit occlusion relationships, which are visually explicit but textually ambiguous.

To mitigate instability in generation, recent efforts (\eg, SVGen~\cite{svgen_wang_2025}, Reason-SVG~\cite{reasonsvg_xing_2025}, and RLRF~\cite{rlrf_rodriguez_2025}) introduce Reinforcement Learning with Verifiable Rewards (RLVR), such as GRPO~\cite{guo2025deepseek}, to optimize large models. By utilizing renderability or visual alignment scores as feedback, these methods guide the model toward better results. However, we argue that scalar rewards essentially \emph{compress} rich visual information into a single number~\cite{yue2025does}, thereby underutilizing the inherent capabilities of MLLMs. Modern MLLMs are equipped with powerful vision encoders designed to perceive dense visual details~\cite{qwen3vl_bai_2025,gpt4_achiam_2023}. Instead of reducing visual feedback to a sparse reward signal, a more direct and information-rich approach is to leverage the vision encoder to explicitly \emph{see} the evolving canvas as \emph{Visual Context}.

Based on this insight, we propose \textbf{Render-in-the-Loop}, a novel generation paradigm (\cref{fig:teaser}). The core idea is intuitive: decomposing the generation process into steps, rendering the partial SVG, and feeding it back to the MLLMs as a visual prompt for the next step. This paradigm is intrinsically aligned with the nature of vector graphics, where visual complexity arises from the sequential layering of primitives. By synchronizing the model's ``drawing hand'' with its ``seeing eye'', this loop allows the model to handle complex occlusion relationships and layer dependencies that are textually implicit but visually obvious.

However, \textbf{simply providing visual context to an off-the-shelf model is insufficient.} Our empirical observations reveal that naively applying this loop to pre-trained MLLMs (\eg, Qwen3-VL~\cite{qwen3vl_bai_2025}, GPT-5~\cite{GPT-5}) yields negligible improvement or even degradation across all evaluation metrics (see~\cref{tab:naive_vsf}). This is because pre-trained models struggle to condition the generation of the next geometric primitive on intermediate visual states.

To bridge this gap, we develop a comprehensive framework encompassing \textbf{data construction, training, and inference verification}. Specifically, we utilize fine-grained path decomposition to construct dense multi-step visual trajectories for the \emph{Visual Self-Feedback (VSF) training}. This explicitly aligns the model's visual perception with the incremental code generation process, enabling the vision encoder to effectively guide precise coordinate prediction based on the current canvas. During inference, leveraging the stepwise structure, we introduce a \emph{Render-and-Verify (RaV)} strategy. Serving as a gatekeeper, RaV detects and filters out redundant or degenerate generation steps that result in negligible visual changes, effectively mitigating degenerate repetitions in generation. 

To the best of our knowledge, the proposed Render-in-the-Loop mechanism represents a novel paradigm for LLM-based SVG generation. Importantly, it is complementary to existing code-domain improvements (\eg, specialized semantic tokens)~\cite{llm4svg_xing_2025,omnisvg_yang_2025,wang2025internsvg}, curriculum learning strategies~\cite{svgen_wang_2025,wang2025internsvg}, and optimization frameworks (\eg, Reinforcement Learning)~\cite{svgen_wang_2025,reasonsvg_xing_2025,rlrf_rodriguez_2025}. In particular, VSF and RL address different aspects of SVG generation: VSF builds a step-wise visual state where the model observes the partial canvas before predicting the next fragment, while RL optimizes the policy with rewards from validity, semantic alignment, or rendered visual fidelity. As these are naturally compatible, Render-in-the-Loop can function both as a standalone enhancement and as a general underlying paradigm that can be combined with complementary optimization techniques, potentially unlocking stronger capabilities for vector generation. We leave a full study of combining VSF with RL to future work.

Our main contributions are summarized as follows:

\begin{itemize}
    \item \textbf{Paradigm Innovation.} We identify the limitations of ``blind'' autoregression and scalar-reward RL in SVG generation. We propose \textbf{Render-in-the-Loop}, which unlocks the visual potential of MLLMs by establishing a direct feedback loop between the rendered canvas and the generation process.
    
    \item \textbf{Synergistic Training and Inference.} We demonstrate that naive visual feedback is ineffective for foundational models. To address this, we introduce the \emph{Visual Self-Feedback (VSF) training} to condition the next primitive generation on intermediate visual states, complemented by an inference-time Render-and-Verify (RaV) mechanism to suppress generation redundancy.
    
    \item \textbf{Strong Performance and Exceptional Data Efficiency.} Built upon the Qwen architectures~\cite{qwen3vl_bai_2025}, our approach achieves highly competitive performance on the standard MMSVGBench~\cite{omnisvg_yang_2025}. Notably, by learning from merely a 0.85M subset of the 2M open-source corpus provided by OmniSVG~\cite{omnisvg_yang_2025}, our model surpasses the performance of both the OmniSVG baseline itself and the recent InternSVG~\cite{wang2025internsvg} trained on 16M samples. This empirical evidence (see~\cref{tab:main_results}) demonstrates that explicit visual feedback is a more critical factor than raw data scaling for synthesizing high-quality vector graphics.
\end{itemize}

    \section{Methodology}

In this section, we present the proposed Render-in-the-Loop paradigm for vector graphic generation. We first introduce the Visual Self-Feedback (VSF) training framework in~\cref{sec:VSF}. We then describe the data preprocessing strategy tailored for constructing the training sequences in \cref{sec:data_preprocessing}. Finally, we introduce a Render-and-Verify (RaV) decoding strategy for real-time error mitigation at inference time in \cref{sec:infer}.

\subsection{Visual Self-Feedback Training Framework}
\label{sec:VSF}

We reformulate SVG generation not as a one-shot translation task (as seen in prior works~\cite{starvector_rodriguez_2025}), but as a \textbf{step-wise drawing process} grounded in visual feedback. In this framework, the model functions as both the ``hand'' (generating XML code) and the ``eye'' (perceiving the canvas), learning to paint incrementally based on the evolving visual state. This sequential modeling aligns seamlessly with the intrinsic nature of vector rendering, where correct visual composition relies on the \textbf{precise ordering of layers}. By observing the intermediate canvas, the model effectively learns to resolve \textbf{complex occlusion relationships}.

\subsubsection{Step-wise Drawing Formulation}

Formally, a complete SVG drawing session is modeled as a unified multimodal sequence containing interleaved user conditions, geometric code segments, and rasterized canvas states. Let $P$ denote the user's prompt, which can be either a textual description (for text-to-SVG) or a reference image (for image-to-SVG). The generation process is decomposed into $N$ steps, where each step $t$ produces a code fragment $C_t$ (\eg, a path). The sequence is structured as:
\begin{equation}
\mathcal{S} = [P, C_1, I_1, C_2, I_2, \dots, C_N, I_N, \langle \text{END} \rangle],
\end{equation}
where \textbf{(1)} $C_t$ is the SVG code segment generated at step $t$; \textbf{(2)} $I_t = R(C_1 \oplus \dots \oplus C_t)$ is the rasterized image of the \emph{cumulative} canvas after applying all code segments up to step $t$. This efficiently captures the evolving occlusion and spatial relationships. The rasterization function $R$ is realized via standard vector graphics rendering pipelines; and \textbf{(3)} $\langle \text{END} \rangle$ is a special token indicating drawing completion. By comparing the intermediate canvas $I_t$ against the target prompt $P$, the model explicitly learns \emph{when to stop} to prevent redundant over-drawing.

Unlike standard pure code generation scenarios~\cite{omnisvg_yang_2025,svgen_wang_2025}, our model conditions its prediction on the \emph{full drawing history}. At any step $t$, to generate the next code segment $C_t$, the model attends to the entire preceding context:
\begin{equation}
X_t = [P, C_1, I_1, \dots, C_{t-1}, I_{t-1}].
\end{equation}
Input images $I_k$ are processed by the vision encoder of Multimodal LLMs~\cite{qwen3vl_bai_2025} and injected as visual tokens into the sequence. This ensures that the model explicitly perceives the valid visual outcome of its previous generated code before drawing the next stroke.

\subsubsection{Training Objective}

We employ standard \textbf{Multiturn Visual Instruction Tuning} to train the model. The training data consists of constructed step-wise drawing sequences (details in \cref{sec:data_preprocessing}). We optimize the model using the standard autoregressive language modeling loss over the entire sequence.

Let $\mathbf{y}$ represent the flattened sequence of tokens for the entire session $\mathcal{S}$. We define a binary mask $M$ where $M_i = 1$ if the token $y_i$ belongs to the model's output (\ie, tokens within any $C_t$ or the $\langle \text{END} \rangle$ token), and $M_i = 0$ otherwise (\ie, for $P$ and all inserted image tokens $I_t$). The VSF loss objective is:
\begin{equation}
\mathcal{L}_{\text{VSF}} = - \frac{1}{\sum_{j=1}^{|\mathbf{y}|} M_j} \sum_{i=1}^{|\mathbf{y}|} M_i \cdot \log P_{\theta}(y_i \mid y_{<i}),
\end{equation}
where $y_{<i}$ denotes all preceding tokens in the sequence. Key characteristics of this objective include: \textbf{(1) Learning ``How to Draw'':} By observing the visual difference between the current canvas $I_{t-1}$ and the goal $P$, the model learns the physical mapping to deduce the correct next stroke $C_t$. \textbf{(2) Learning ``When to Stop'':} The loss penalizes missing or premature $\langle \text{END} \rangle$ tokens, ensuring the model halts exactly when the intermediate canvas aligns with the target. \textbf{(3) Retaining Historical Context:} Rather than training on isolated state-to-action pairs, jointly optimizing the entire sequence allows the model to remember its past drawings. This ensures the model not only reacts to the current canvas but also follows a coherent, continuous drawing logic to complete the graphic.

\subsection{Data Preprocessing for Step-wise Drawing}
\label{sec:data_preprocessing}

To enable the MLLM to \emph{observe} intermediate rendering results during generation, we transform conventional static \emph{Text/Image-to-Full-Code} datasets into interleaved multi-step image-text trajectories.

We adopt the open-source dataset subset from OmniSVG~\cite{omnisvg_yang_2025} as our primary data source. While the original work reports training on 2 million samples, only a portion was released (0.9M icons and 0.25M illustrations). To ensure data quality, we perform strict deduplication by filtering out samples with identical code strings, resulting in a refined set of 0.65M icons and 0.2M illustrations. Despite using this smaller, cleaner subset, our method demonstrates superior performance (see \cref{sec:experiments}), highlighting the effectiveness of our visual feedback mechanism over raw data scale.

The original SVG files are often optimized for storage, containing an average of only about 4 \texttt{<path>} elements per file. Simply splitting generation by these original elements would yield sparse visual states. Furthermore, generating a single complex path with numerous coordinates in one go challenges autoregressive models. We therefore perform fine-grained decomposition of lengthy paths to increase the density of visual feedback while maintaining rendering fidelity.

\subsubsection{Fine-grained Path Decomposition}
\begin{figure}[t]
    \centering
    \includegraphics[width=\textwidth]{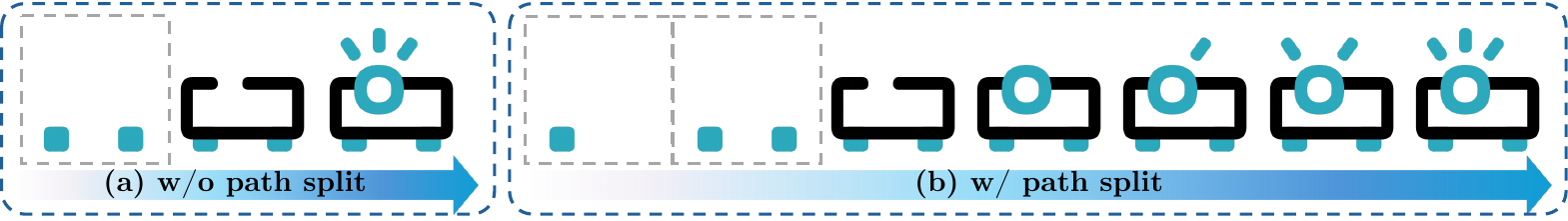}
    \caption{\textbf{Illustration of Fine-grained Path Decomposition.} We compare the intermediate visual states without (a) and with (b) our path decomposition. Without splitting, a complex original path containing multiple visually disjoint elements (\eg, projector body, lens, beams) is rendered abruptly in a single step, yielding sparse visual states. In contrast, our pipeline geometrically decomposes it into atomic subpaths, ensuring each distinct element is drawn sequentially. This significantly increases semantic feedback density, allowing the model to progressively learn spatial details.}
    \label{fig:path_split}
\end{figure}

Naively truncating the \texttt{d} attribute of a long \texttt{<path>} into multiple shorter paths often breaks SVG rendering semantics (\eg, fill rules, subpath closure). To address this, we design a geometric decomposition pipeline. \cref{fig:path_split} illustrates an example of the decomposed results. The pipeline consists of three steps: \textbf{(1) Subpath Extraction:} We parse the drawing commands in the \texttt{d} attribute and decompose segments starting from non-contiguous commands (\eg, \texttt{M}/\texttt{MoveTo}) into independent subpaths. \textbf{(2) Topology Analysis:} We treat each subpath as a 2D polygon and construct a dependency graph. An edge is added if subpaths have a containment relationship (often defining graphical holes via fill rules) or if they intersect spatially with non-unity opacity (to avoid color blending artifacts upon overlapping). \textbf{(3) Connected Component Merging:} We merge subpaths within the same connected component of the dependency graph back into a single \texttt{<path>} element. This ensures visually coupled elements remain atomic while independent elements are separated to provide more frequent visual feedback steps.

\subsubsection{Curriculum Construction}

After fine-grained decomposition, the average number of \texttt{<path>} elements per SVG increases from 4 to 6. This preprocessing significantly improves the \emph{density of visual intermediate states}. It also introduces an implicit curriculum: instead of drawing complex composite shapes in a single attempt, the model learns to sequentially compose simpler geometric primitives.

Finally, we convert each decomposed SVG into the standard interleaved format described in \cref{sec:VSF}. The sequence $\mathcal{S}$ is constructed by rendering the cumulative canvas after each step $t$ to obtain $I_t$, which is then inserted as the visual context for the subsequent step $C_{t+1}$.

\subsection{Render-and-Verify}
\label{sec:infer}

Standard autoregressive models often lack explicit mechanisms to correct errors during generation, leading to issues such as \emph{repetitive loops}, \emph{off-canvas drawing}, and \emph{redundant over-drawing}. In our framework, since the model is conditioned on the rendered history, we introduce a lightweight \textbf{Render-and-Verify (RaV)} strategy at inference time to detect and mitigate these degenerate behaviors without additional training. At inference step $t$, the model generates a candidate code segment $\hat{C}_t$. Before accepting it, we render a hypothetical future canvas state $\hat{I}_t = R(C_{1:t-1} \oplus \hat{C}_t)$ and apply two heuristic checks. \textbf{Visual Difference Check:} We compare $\hat{I}_t$ with the previous state $I_{t-1}$. If the pixel-wise difference is below a threshold $\epsilon$ (indicating no visual contribution), $\hat{C}_t$ is rejected. \textbf{Repetition Check:} If the string similarity between $\hat{C}_t$ and $C_{t-1}$ exceeds a threshold $\tau_{\text{sim}}$, $\hat{C}_t$ is also rejected to prevent repetitive loops. 

If the candidate $\hat{C}_t$ fails to pass either of these two heuristic checks, it indicates that the proposed code segment is either visually stagnant or structurally redundant. In such cases, we discard $\hat{C}_t$ and trigger an adaptive resampling mechanism. Specifically, we adjust the generation hyperparameters (\eg, slightly increasing the sampling temperature) to encourage the model to escape local repetitive distributions and explore alternative, valid drawing sequences. This resampling is continuously repeated until a verified step is found. We emphasize that RaV is \emph{not} the primary stopping mechanism: during VSF training, the model is explicitly supervised to emit the $\langle \text{END} \rangle$ token when the rendered canvas matches the target, so under normal generation the model itself decides when to stop. RaV only acts as an inference-time guardrail that rejects stagnant or near-duplicate candidates and resamples adaptively. Forced termination is a last-resort protection that is triggered only after repeated failures: if the model consistently fails to produce a valid segment after a predefined maximum number of retries ($K_{\max}$), the policy is typically trapped in a repetitive local distribution or lacks the capability to complete the remaining content, so continuing would mostly add noisy paths. In this case we force the current step to output the $\langle \text{END} \rangle$ token to prevent the generation process from stalling. Overall, this Verify-then-Accept strategy functions as a lightweight yet highly effective inference-time guardrail. It dynamically filters out invalid generation steps on the fly, striving to ensure that every executed drawing action provides a concrete and meaningful visual contribution to the final graphic.

    \section{Experiments}
\label{sec:experiments}

To comprehensively evaluate the proposed Render-in-the-Loop paradigm, our evaluation focuses on three key questions:
\textbf{(Q1)} Does the explicit visual feedback mechanism enable our model to outperform existing strong open-weight baselines, including larger-scale models (\eg, OmniSVG and InternSVG) and RL-enhanced methods (\eg, SVGen)?
\textbf{(Q2)} How much does the Visual Self-Feedback (VSF) training strategy contribute to the model's generation quality compared to standard Supervised Fine-Tuning (SFT)?
\textbf{(Q3)} Can the inference-time Render-and-Verify (RaV) strategy effectively mitigate degenerate repetitions in the generation process?

\subsection{Experimental Setup}

\paragraph{Datasets and Benchmarks.} Following OmniSVG~\cite{omnisvg_yang_2025}, we use the \textbf{MMSVG} dataset. Although OmniSVG was trained on 2M samples, only a subset is publicly released; we train strictly on this subset (\textbf{0.65M icons and 0.2M illustrations}), \ie, less than 50\% of the baseline's data. We apply our Fine-grained Path Decomposition (\cref{sec:data_preprocessing}) to organize the data into in-the-loop visual trajectories, and evaluate on the official \textbf{MMSVGBench}~\cite{omnisvg_yang_2025}.

\paragraph{Baselines.} We compare our method against three categories of strong baselines: (1) \textbf{Optimization-based Methods:} DiffVG~\cite{diffvg_li_2020}, LIVE~\cite{live_ma_2022}, VectorFusion~\cite{vectorfusion_jain_2023} and SVGDreamer~\cite{svgdreamer_xing_2024}; (2) \textbf{LLM/VLM-based Methods:} StarVector~\cite{starvector_rodriguez_2025} (8B), IconShop~\cite{iconshop_wu_2023}, GPT-5~\cite{GPT-5}, OmniSVG~\cite{omnisvg_yang_2025} (8B, NeurIPS 2025), and InternSVG~\cite{wang2025internsvg} (8B, ICLR 2026). Note that InternSVG is trained on a massive dataset of 16M samples (SAgoge). We ensure that the primary open-source baselines are of comparable model size to ours (8B); (3) \textbf{RL-based Methods:} SVGen~\cite{svgen_wang_2025} (7B). As the only open-sourced method currently applying Reinforcement Learning (specifically GRPO) to SVG generation, we select it as the representative baseline for this category.

\paragraph{Evaluation Metrics.} We employ a comprehensive set of metrics to assess both visual quality and semantic alignment. For \textbf{Text-to-SVG}, we report FID (Fréchet Inception Distance)~\cite{heusel2017gans} to measure distribution-level image quality and CLIP Score~\cite{clip_radford_2021} to evaluate semantic consistency with the prompt. We also report Aesthetic Score and Human Preference Score~\cite{wu2023human}. For \textbf{Image-to-SVG}, we measure reconstruction fidelity using MSE, LPIPS~\cite{lpips_zhang_2018}, SSIM~\cite{ssim_wang_2004}, and DINO Score~\cite{oquab2023dinov2}.

\paragraph{Implementation Details.} Our model is initialized from Qwen3-VL-8B-Instruct~\cite{qwen3vl_bai_2025} and fine-tuned for 3 epochs on 8 NVIDIA H100 GPUs (AdamW, learning rate $1e-5$, cosine decay). All intermediate SVGs are rendered at $224 \times 224$ and encoded by the Vision Transformer~\cite{dosovitskiy2020image} into only 49 visual tokens, so that interleaving multiple canvases does not overwhelm the context window (maximum sequence length $10240$). During inference, we employ RaV with $\epsilon=0.001$ and $\tau_{\text{sim}}=0.98$.

\begin{table}[t]
\centering
\caption{Quantitative results between our method and current strong text-to-SVG and image-to-SVG baselines on MMSVG benchmarks. Our model demonstrates superior SVG generation performance. Best results are in \textbf{bold}, and second best are \underline{underlined}.}
\label{tab:main_results}
\resizebox{\textwidth}{!}{%
\begin{tabular}{llcccccccc}
\toprule
\multirow{2}{*}{\textbf{Dataset}} & \multirow{2}{*}{\textbf{Methods}} &
\multicolumn{4}{c}{\textbf{Text-to-SVG}} &
\multicolumn{4}{c}{\textbf{Image-to-SVG}} \\
\cmidrule(lr){3-6} \cmidrule(lr){7-10}
& & FID$\downarrow$ & CLIP$\uparrow$ & Aes$\uparrow$ & HPS$\uparrow$ &
DINO$\uparrow$ & SSIM$\uparrow$ & LPIPS$\downarrow$ & MSE$\downarrow$ \\
\midrule

\multirow{13}{*}{\textbf{MMSVG-Icon}}
& \multicolumn{9}{c}{\textit{Optimization-based Methods}} \\
\cmidrule(lr){2-10}
& VectorFusion & \textbf{250.77} & \textbf{0.240} & \textbf{4.76} & \textbf{0.237} & - & - & - & - \\
& SVGDreamer   & 308.94 & 0.207 & 4.26 & 0.221 & - & - & - & - \\
& LIVE         & - & - & - & - & 0.932 & 0.943 & 0.106 & 0.011 \\
& DiffVG       & - & - & - & - & \textbf{0.940} & \textbf{0.954} & \textbf{0.066} & \textbf{0.002} \\
\addlinespace
\cmidrule(lr){2-10}

& \multicolumn{9}{c}{\textit{Autoregressive Methods}} \\
\cmidrule(lr){2-10}\
& StarVector(8B)  & - & - & - & - & 0.895 & 0.881 & 0.231 & 0.059 \\
& GPT-5           & - & - & - & - & 0.902 & 0.825 & 0.387 & 0.113 \\
& IconShop        & 213.28 & 0.288 & 4.55 & 0.244 & - & - & - & - \\
& SVGen(RL-based) & 129.23 & 0.281 & 4.78 & 0.244 & - & - & - & - \\
& OmniSVG(8B)     & 130.56 & 0.276 & 4.60 & 0.242 & 0.922 & 0.893 & 0.235 & 0.040 \\
& InternSVG(8B) & 128.80 & 0.291 & 4.75 & 0.246 & 0.926 & 0.901 & \underline{0.182} & 0.034 \\
& Qwen3-VL (SFT only) & 148.72 & 0.262 & 4.41 & 0.232 & 0.889 & 0.842 & 0.312 & 0.094 \\
& \textbf{Ours (VSF)} & \underline{127.98} & \textbf{0.294} & \underline{4.80} & \underline{0.249} &
\underline{0.928} & \underline{0.908} & 0.188 & \underline{0.031} \\
& \textbf{Ours (VSF+RaV)} & \textbf{127.64} & \underline{0.293} & \textbf{4.86} & \textbf{0.251} &
\textbf{0.931} & \textbf{0.914} & \textbf{0.172} & \textbf{0.027} \\
\midrule

\multirow{13}{*}{\textbf{MMSVG-Illustration}}
& \multicolumn{9}{c}{\textit{Optimization-based Methods}} \\
\cmidrule(lr){2-10}
& VectorFusion & \textbf{253.94} & 0.185 & \textbf{4.94} & \textbf{0.226} & - & - & - & - \\
& SVGDreamer   & 419.70 & \textbf{0.201} & 4.37 & 0.221 & - & - & - & - \\
& LIVE         & - & - & - & - & 0.935 & 0.950 & 0.111 & 0.008 \\
& DiffVG       & - & - & - & - & \textbf{0.945} & \textbf{0.955} & \textbf{0.065} & \textbf{0.001} \\
\addlinespace
\cmidrule(lr){2-10}

& \multicolumn{9}{c}{\textit{Autoregressive Methods}} \\
\cmidrule(lr){2-10}
& StarVector(8B)  & - & - & - & - & 0.877 & 0.900 & 0.238 & 0.046 \\
& GPT-5           & - & - & - & - & 0.894 & 0.876 & 0.345 & 0.064 \\
& IconShop        & 137.93 & 0.233 & 4.46 & 0.224 & - & - & - & - \\
& SVGen(RL-based) & 139.53 & 0.220 & 4.61 & 0.230 & - & - & - & - \\
& OmniSVG(8B)     & 138.42 & 0.231 & 4.51 & 0.232 & 0.905 & 0.907 & 0.231 & 0.031 \\
& InternSVG(8B) & 138.10 & 0.229 & 4.58 & 0.234 & 0.912 & 0.915 & 0.205 & 0.026 \\
& Qwen3-VL (SFT only) & 155.84 & 0.221 & 4.33 & 0.221 & 0.882 & 0.861 & 0.318 & 0.082 \\
& \textbf{Ours (VSF)} & \underline{137.86} & \underline{0.235} & \underline{4.69} & \underline{0.241} &
\underline{0.913} & \underline{0.921} & \underline{0.193} & \underline{0.024} \\
& \textbf{Ours (VSF+RaV)} & \textbf{137.79} & \textbf{0.237} & \textbf{4.73} & \textbf{0.242} &
\textbf{0.918} & \textbf{0.928} & \textbf{0.178} & \textbf{0.021} \\
\bottomrule
\end{tabular}%
}
\end{table}
\subsection{Comparison}

\subsubsection{Quantitative Comparison}
\label{sec:quantitative}
We present the quantitative results on both Text-to-SVG and Image-to-SVG tasks in~\cref{tab:main_results}. Our Render-in-the-Loop model achieves highly competitive performance on both Icon and Illustration subsets.

\paragraph{Text-to-SVG Analysis.}
As shown in~\cref{tab:main_results} (Left), our method significantly outperforms the optimization-based baselines (VectorFusion, SVGDreamer) in terms of editability and generation quality. More importantly, compared to \textbf{OmniSVG} (trained on 2M data), our method (trained on 0.85M) achieves a lower FID and higher CLIP Score. 
Remarkably, our method even surpasses the recent open-weight leader \textbf{InternSVG} (trained on a massive \textbf{16M} dataset (approx. $20\times$ larger than ours)). For instance, on the Icon subset, we achieve a lower FID (127.64 vs. 128.80) and higher CLIP Score (0.293 vs. 0.291).
This verifies our core hypothesis: \emph{seeing} the clear rendering process and explicit layer-wise dependencies is far more effective than \emph{blindly} learning from massive text-code pairs. Furthermore, our method outperforms the RL-based \textbf{SVGen}, suggesting that dense visual feedback provides richness that scalar rewards cannot match.

\paragraph{Image-to-SVG Analysis.}
\cref{tab:main_results} (Right) summarizes the reconstruction results. Our method shows superior reconstruction, especially on the complex Illustration subset: the high DINO score reflects accurate semantic structure, and conditioning on intermediate renderings yields better structural consistency and significantly lower LPIPS than the open-loop OmniSVG.

\begin{figure}[t]
    \centering
    \includegraphics[width=\linewidth]{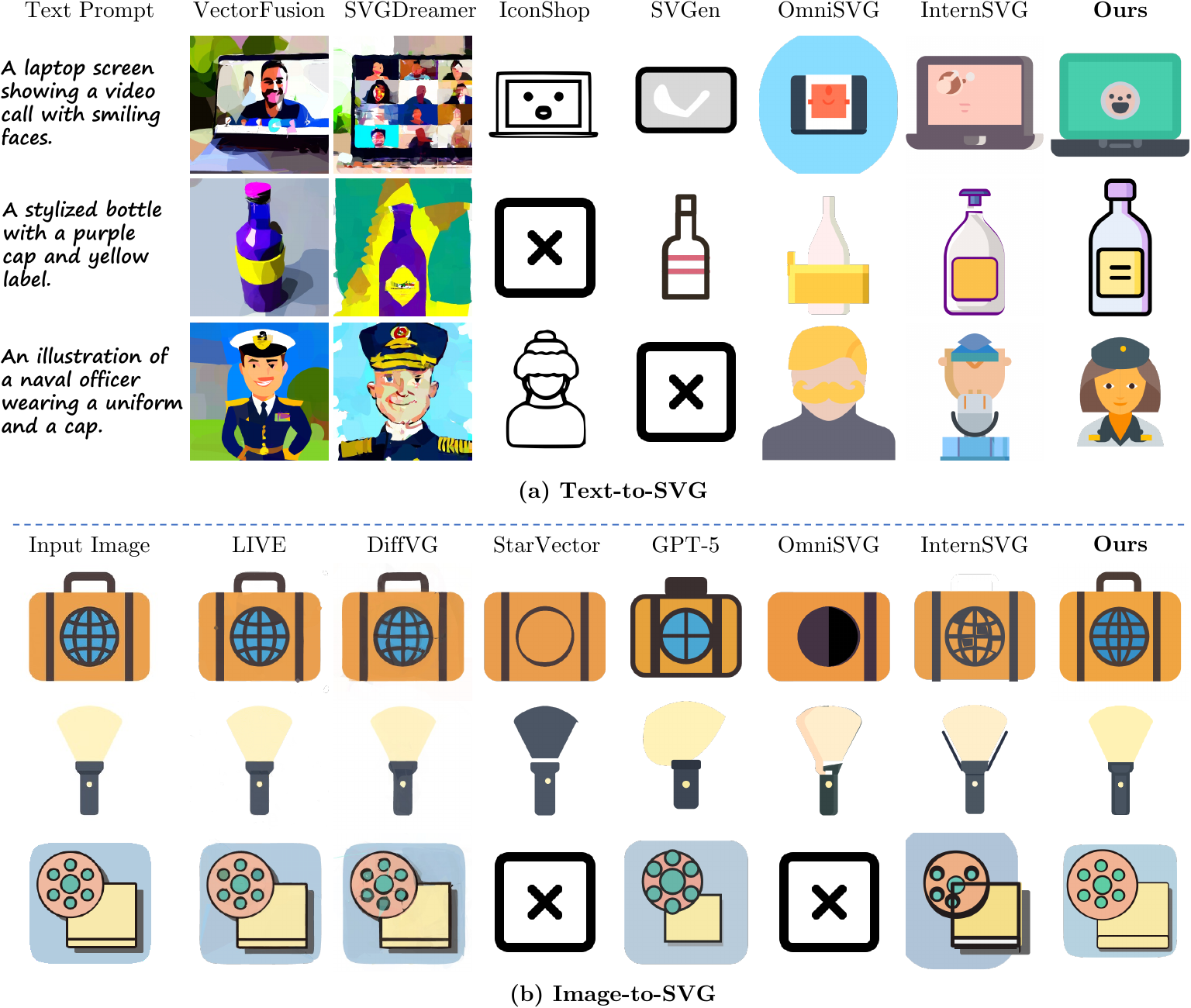}
    \caption{Qualitative comparison on MMSVG benchmarks. (a) In Text-to-SVG, our method generates clear semantic details (\eg, facial expressions, uniform details) where baselines like OmniSVG often fail or produce noise. (b) In Image-to-SVG, our model effectively reconstructs internal structures (\eg, flashlight lens, briefcase patterns), demonstrating superior fidelity to the input raster image.}
    \label{fig:qualitative}
\end{figure}
\subsubsection{Qualitative Comparison}
\cref{fig:qualitative} compares our Render-in-the-Loop against representative optimization-based (\eg, VectorFusion~\cite{vectorfusion_jain_2023}, DiffVG~\cite{diffvg_li_2020}) and autoregressive (\eg, OmniSVG~\cite{omnisvg_yang_2025}, InternSVG~\cite{wang2025internsvg}) baselines.
In Text-to-SVG (\cref{fig:qualitative}a), optimization-based methods produce messy, overlapping paths with poor editability (\eg, the chaotic background of the ``naval officer''), while autoregressive baselines often yield distorted details (\eg, glitched faces in the ``laptop screen''). In contrast, our method generates compact, topologically clean primitives with superior semantic alignment.
In Image-to-SVG (\cref{fig:qualitative}b), DiffVG and LIVE overfit the raster image with thousands of tiny paths, sacrificing editability, whereas blind autoregressive models miss internal structures (\eg, the ``flashlight'' lens). Our model faithfully reconstructs geometry while keeping a clean, editable result suitable for design workflows.

\paragraph{Fine-grained Instruction Following.}
Our model also exhibits strong fine-grained instruction-following ability. For prompts such as ``a pink camera with blue accents'' and ``a blue t-shirt with two yellow buttons'', the generated SVGs preserve the requested attribute–object bindings: the model places each attribute on the intended part and keeps this assignment stable throughout the drawing process. We provide qualitative instruction-following examples, together with an extended gallery of diverse generations that demonstrates our model's versatility across a wide variety of subjects, in the supplementary material.

\subsection{Ablation Studies}
\label{sec:ablation}
To investigate the effectiveness of each component in our framework, we conduct ablation studies by comparing our full model (VSF + RaV) with two variants: the base model fine-tuned using standard SFT (Qwen3-VL SFT only) and the model trained with VSF but without inference-time verification (Ours w/o RaV). Detailed results are presented in~\cref{tab:main_results}.

\paragraph{Necessity of VSF Training.}
As argued in our introduction, simply providing visual context (an intermediate canvas) to an off-the-shelf model without specialized training is insufficient. To empirically validate this, we design a \emph{naive multi-turn prompting} baseline: we prompt standard foundational models to perform in-the-loop generation by observing intermediate rendering results across turns, without any VSF training. We stress that this baseline is intended to probe inference-time multi-turn prompting with rendered feedback, \emph{not} to compare foundation models; we deliberately include both an open-source (Qwen3-VL) and a closed-source (GPT-5) MLLM, and evaluate them on both Text-to-SVG and Image-to-SVG to isolate the effect of the prompting protocol from the choice of backbone.
As reported in~\cref{tab:naive_vsf}, the naive multi-turn prompting formulation brings no consistent gain and often degrades performance (especially Aesthetic and HPS) compared to standard one-shot open-loop generation. For instance, GPT-5's LPIPS notably increases from 0.345 to 0.388 on the Illustration subset. This consistent trend across both backbones and tasks confirms that intermediate canvases alone are insufficient: pre-trained models intuitively struggle to align intermediate visual states with the incremental geometry code space. Therefore, our VSF data trajectory construction and training strategy are necessary to unlock the capabilities of the Render-in-the-Loop paradigm.
\begin{figure}[t]
    \centering
    \includegraphics[width=\linewidth]{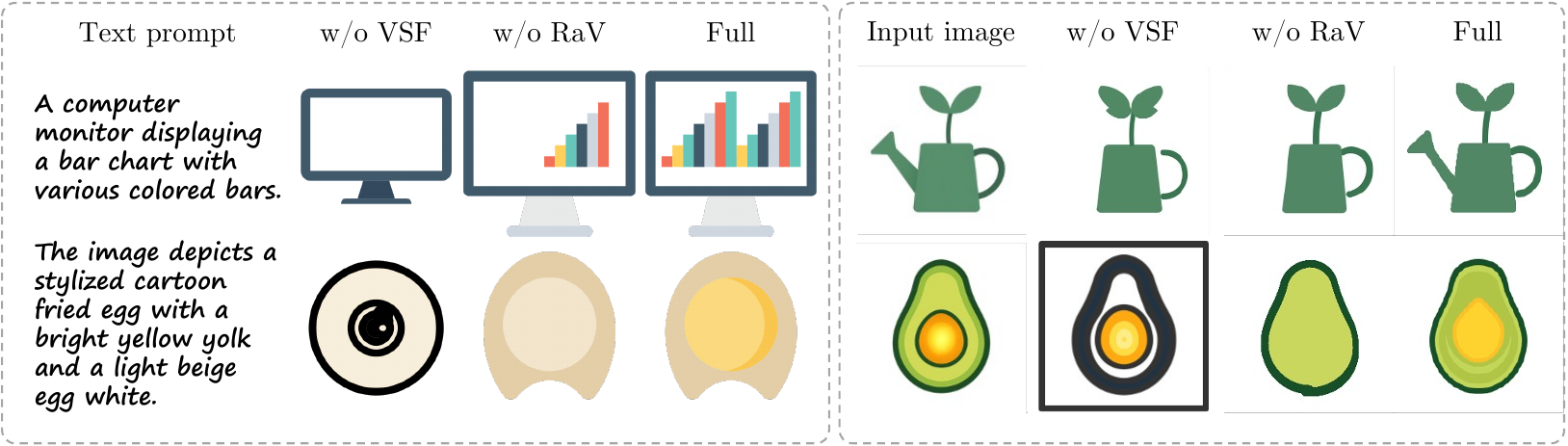}
    \caption{Qualitative ablation study. \textbf{w/o VSF} fails to capture semantic details (\eg, missing bar chart, hallucinating an eye for an egg). \textbf{w/o RaV} suffers from degenerate repetition (\eg, re-drawing the same leaf), leading to incomplete shapes. Our full model generates correct and complete graphics.}
    \label{fig:ablation}
\end{figure}
\begin{table}[t]
\centering
\caption{Effect of \emph{naive multi-turn prompting}. Directly feeding intermediate visual canvases into standard models (without VSF training) brings no consistent gain and often degrades performance across both Text-to-SVG and Image-to-SVG tasks, for both open-source (Qwen3-VL) and closed-source (GPT-5) backbones.}
\label{tab:naive_vsf}

\resizebox{\linewidth}{!}{
\begin{tabular}{llcccccccc}
\toprule

\textbf{Dataset} & \textbf{Setting} 
& \multicolumn{4}{c}{\textbf{Text-to-SVG}} 
& \multicolumn{4}{c}{\textbf{Image-to-SVG}} \\

\cmidrule(lr){3-6} \cmidrule(lr){7-10}

& 
& CLIP$\uparrow$ & Aes$\uparrow$ & HPS$\uparrow$ & FID$\downarrow$
& DINO$\uparrow$ & SSIM$\uparrow$ & LPIPS$\downarrow$ & MSE$\downarrow$ \\

\midrule

\multirow{2}{*}{\textbf{MMSVG-Icon}}
& One-shot (Open-loop)
& \textbf{0.311} & \textbf{5.026} & \textbf{0.219} & -
& \textbf{0.902} & \textbf{0.825} & \textbf{0.387} & \textbf{0.113} \\

& Naive multi-turn prompting
& 0.294 & 4.925 & 0.185 & -
& 0.891 & 0.814 & 0.405 & 0.121 \\

\midrule

\multirow{2}{*}{\textbf{MMSVG-Illust.}}
& One-shot (Open-loop)
& \textbf{0.293} & \textbf{5.195} & \textbf{0.210} & -
& \textbf{0.894} & \textbf{0.876} & \textbf{0.345} & \textbf{0.064} \\

& Naive multi-turn prompting
& 0.281 & 4.807 & 0.167 & -
& 0.883 & 0.852 & 0.388 & 0.075 \\

\bottomrule
\end{tabular}
}
\end{table}

\paragraph{Effect of Visual Self-Feedback (VSF).} 
As illustrated in~\cref{fig:ablation} (Left), standard SFT models (w/o VSF) suffer from severe hallucinations when handling complex prompts. 
For the ``monitor with bar chart'' prompt, the blind model fails to draw the bars entirely, outputting a generic monitor shape.
Similarly, for the ``fried egg'' prompt, it generates an eye-like structure, misinterpreting the geometric composition. 
By incorporating VSF, the model perceives the evolving canvas, enabling it to correct these semantic deviations and generate identifiable objects. Quantitatively, this corresponds to the significant FID reduction (155.84 $\to$ 137.86) observed in~\cref{tab:main_results}.

\paragraph{Effect of Render-and-Verify (RaV).}
\cref{fig:ablation} (Right) highlights the necessity of the RaV strategy.
Without RaV, the model tends to fall into degenerate repetition loops, continuously generating redundant strokes that do not contribute to the visual progress (\eg, re-drawing the same leaf contour in the ``plant'' example). This leads to incomplete SVGs.
RaV acts as a gatekeeper: by rejecting paths that yield negligible visual differences (Pixel Difference $<\epsilon$), it forces the model to break out of these loops and proceed to draw the next meaningful component (\eg, the watering can body). This ensures both the completeness and conciseness of the final output.

\paragraph{Data Efficiency.}
Despite training on only \textbf{0.85M} samples, our model matches or exceeds baselines trained on far larger datasets (OmniSVG, 2M; InternSVG, 16M), \ie, using less than 6\% of the leading baseline's data. This suggests that the bottleneck in SVG generation is not data scale alone, but the lack of \emph{effective usage} of visual priors.

\paragraph{Robustness to Occlusion.}
Vector graphics often involve complex layer stacking (\eg, background before foreground). Open-loop models frequently mis-order layers, producing ``hidden'' geometries that waste tokens. By observing the canvas, our model naturally learns to draw back-to-front, resolving occlusion ambiguities that text alone cannot capture.

\subsection{Inference Cost Analysis}
\label{sec:cost}
Since Render-in-the-Loop renders intermediate canvases and interleaves them into the context, it incurs additional inference cost compared to a single-pass open-loop model. To quantify this overhead, we compare our model against an open-loop SFT variant using the same Qwen3-VL backbone, the same vLLM serving stack, and identical NVIDIA V100 GPUs on MMSVGBench. As reported in~\cref{tab:cost}, VSF introduces a $1.50\times$ latency overhead, which closely tracks the $1.46\times$ increase in generated tokens. Notably, the SVG rendering itself is negligible ($0.038$\,s in total), so the overhead originates almost entirely from step-wise autoregressive decoding rather than rasterization. This indicates that the cost of Render-in-the-Loop scales with the number of generated tokens and can be further reduced by standard decoding-acceleration techniques. We further show that RaV is robust to its threshold settings in the supplementary material.

\begin{table}[t]
\centering
\caption{Inference cost on MMSVGBench. The overhead of VSF closely matches the increase in generated tokens, while the rendering cost is negligible.}
\label{tab:cost}
\setlength{\tabcolsep}{5pt}
\resizebox{\linewidth}{!}{
\begin{tabular}{lcccccc}
\toprule
Method & Time(s) & Time Cost & Tokens & Token Cost & LLM(s) & Render(s) \\
\midrule
Open-loop SFT & 28.77 & 1.00$\times$ & 550.94 & 1.00$\times$ & 28.77 & 0.000 \\
Ours (w. VSF) & 43.19 & \textbf{1.50$\times$} & 804.31 & \textbf{1.46$\times$} & 43.14 & \textbf{0.038} \\
\bottomrule
\end{tabular}}
\end{table}

\subsection{Failure Case Analysis}
\label{sec:failure}
\begin{figure}[t]
    \centering
    \includegraphics[width=\linewidth]{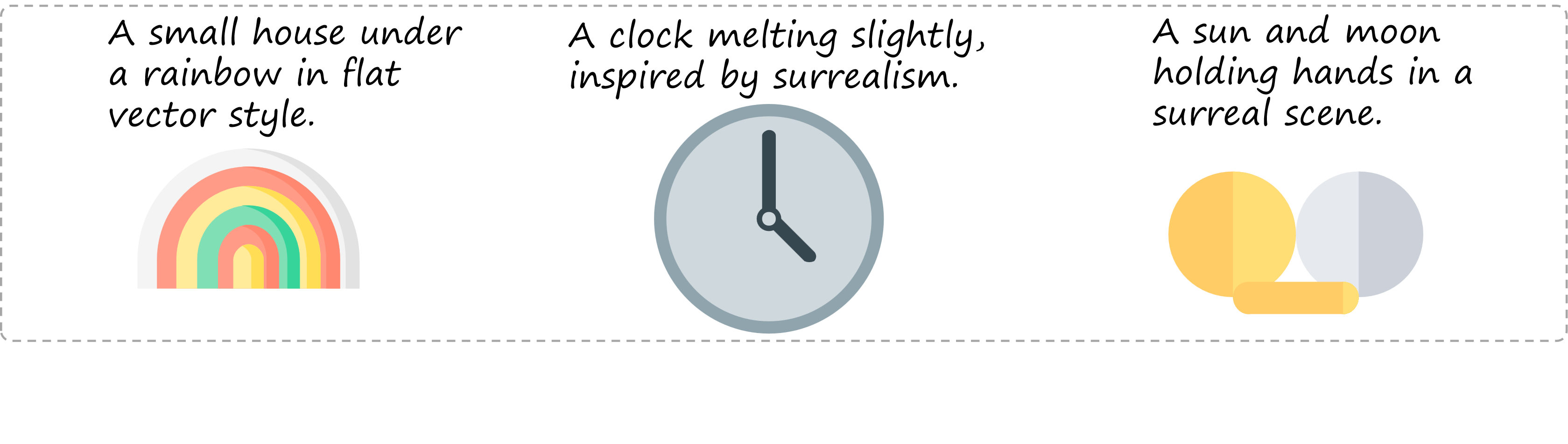}
    \caption{Representative failure cases. From left to right: (a) degenerative repetition (the rainbow is drawn but the house is missing); (b) limited semantic adherence for complex states (a normal clock instead of a melting one); and (c) oversimplification in complex scenes (only a coarse composition is produced).}
    \label{fig:failures}
\end{figure}

While our method achieves strong overall performance, we observe three representative failure modes, illustrated in~\cref{fig:failures}. \textbf{(1) Degenerative repetition:} the model occasionally gets stuck in local repetitive patterns and fails to complete the intended content (\eg, drawing the rainbow but missing the house). \textbf{(2) Limited semantic adherence for complex states:} for prompts describing unusual states or surreal compositions, the model may revert to a generic object (\eg, producing a normal clock instead of a melting clock). \textbf{(3) Oversimplification in complex scenes:} for highly compositional prompts, the model sometimes generates only a coarse layout and omits important details. These cases suggest that, although Render-in-the-Loop substantially improves geometric grounding, faithfully rendering abstract states and richly compositional scenes remains an open challenge, which we discuss further in~\cref{sec:conclusion}.

    \section{Conclusion}
\label{sec:conclusion}

In this paper, we propose \textit{Render-in-the-Loop}, a novel generation paradigm that reformulates SVG generation from an open-loop textual process into an integrated, context-aware visual sequence. By explicitly conditioning the MLLM on its own incremental intermediate rendering results, our framework effectively unlocks the model's visual priors, better handling structural inaccuracies and complex occlusion relationships. Equipped with Visual Self-Feedback training and the Render-and-Verify inference mechanism, our model achieves highly competitive capabilities and exceptional data efficiency on standard benchmarks.

\paragraph{Limitations and Future Work.} 
A current limitation of our approach is the inference overhead. Since our method requires repeatedly rendering intermediate SVG paths and interleaving the resulting images into the context window, the actual inference time is increased compared to single-pass open-loop generation (a $1.50\times$ latency overhead, \cf~\cref{sec:cost}). Furthermore, the intermediate canvases are fixed at $224 \times 224$; the trade-off between perceiving finer details at higher resolutions and computational cost remains unexplored. Another limitation is that our current formulation is purely \emph{additive}: each step only appends new primitives to the canvas, so the model cannot explicitly undo, delete, or replace already-accepted primitives. As a result, errors introduced in early steps can only be visually compensated by later additions rather than corrected. A promising direction is to extend the action space with DOM-level editing operations (\eg, \texttt{DELETE}, \texttt{REPLACE}, and \texttt{MODIFY}), enabling genuine self-correction on top of the Render-in-the-Loop foundation. 
For future work, we plan to further train the model to develop stronger capabilities of self-correction, reasoning, and planning on top of the Render-in-the-Loop foundation, potentially incorporating dynamic multi-scale visual feedback. Beyond SVGs, we believe the core idea of closing the rendering feedback loop can seamlessly generalize to broader inverse rendering code generation tasks, including HTML/CSS for web development, LaTeX/TikZ for scientific visualization, 3D rendering programs, and complex CAD modeling systems. we truly hope this exciting line of work can serve as a highly versatile and general framework for advancing structured, code-driven visual synthesis.

    \section*{Acknowledgements}
This work was supported in part by National Natural Science Foundation of China (No.62572039, No.62461160331, No.62132001). This work was also supported by the NSFC/RGC Collaborative Research Scheme (CRS\_HKU703/24). Dr. Xu's research work described in this paper was conducted in the JC STEM Lab of Multimedia and Machine Learning funded by The Hong Kong Jockey Club Charities Trust.

    \bibliographystyle{splncs04}
    \bibliography{main}

    \clearpage
    \renewcommand{\thefigure}{S\arabic{figure}}
\setcounter{figure}{0}
\renewcommand{\thetable}{S\arabic{table}}
\setcounter{table}{0}
\renewcommand{\theequation}{S\arabic{equation}}
\setcounter{equation}{0}

\appendix


\section*{Overview}
In this supplementary material, we provide additional details related to our work on \textbf{Render-in-the-loop}. Specifically, this document covers the following aspects:

\begin{itemize}[left=0pt]
    \item \textbf{Additional Experiments and Analyses} (\cref{sec:supp_analyses}): Provides a sensitivity study of the RaV thresholds and additional qualitative results.
    \item \textbf{Implementation Details} (\cref{sec:supp_impl_details}): Provides extended details on our Render-and-Verify filtering strategy during inference, the system prompts for our models, and concrete examples of the visual self-feedback training format.
    \item \textbf{More Details of the Baselines} (\cref{sec:morebaselines}): Provides expanded overviews of the various baseline models included in our comparative study.
    \item \textbf{Comparison with LIVE and DiffVG} (\cref{sec:showcase_baselines}): Offers a more granular visual critique of the stroke-stacking problem prevalent in optimization-based methods compared to our structured generation.
    \item \textbf{MMSVGBench} (\cref{sec:mmsvgbench}): Details the structure and metrics of the multimodal benchmark employed in our capability evaluations.
    \item \textbf{Related Work} (\cref{sec:supp_related}): Reviews optimization-based, autoregressive, and feedback-driven SVG generation, and positions our Render-in-the-Loop paradigm with respect to prior work.

\end{itemize}

\section{Additional Experiments and Analyses}
\label{sec:supp_analyses}
This section provides supplementary analyses that complement the main paper, covering the robustness of the Render-and-Verify (RaV) thresholds and additional qualitative results.

\subsection{Robustness of RaV Thresholds}
\label{sec:supp_rav}
RaV is designed to filter visually non-contributive or near-duplicate fragments, rather than to optimize semantic quality, so its behavior should not be sensitive to the exact threshold values. To verify this, we vary the visual-difference threshold $\epsilon$ and the repetition threshold $\tau_{\text{sim}}$ around the default setting on MMSVG-Illustration Text-to-SVG. As shown in~\cref{tab:rav}, performance remains stable across \textit{Loose}, \textit{Default}, and \textit{Strict} configurations (CLIP varies within $0.001$, HPS within $0.003$). This confirms that RaV is a lightweight degeneracy filter rather than a brittle, benchmark-specific module that requires careful tuning.

\begin{table}[h]
\centering
\caption{Sensitivity of RaV thresholds on MMSVG-Illustration Text-to-SVG. Performance is stable across loose-to-strict configurations, indicating that RaV is a robust degeneracy filter rather than a finely tuned module.}
\label{tab:rav}
\setlength{\tabcolsep}{6pt}
\renewcommand{\arraystretch}{1.0}
\begin{tabular}{lccccc}
\toprule
Setting & $\epsilon$ & $\tau_{\text{sim}}$ & CLIP$\uparrow$ & Aes.$\uparrow$ & HPS$\uparrow$ \\
\midrule
Loose   & 0.0005 & 0.995 & 0.236 & \textbf{4.76} & 0.241 \\
Default & 0.0010 & 0.980 & \textbf{0.237} & 4.73 & \textbf{0.242} \\
Strict  & 0.0020 & 0.950 & \textbf{0.237} & 4.71 & 0.239 \\
\bottomrule
\end{tabular}
\end{table}

\subsection{Additional Qualitative Results}
\label{sec:supp_qualitative}
\begin{figure}[h]
    \centering
    \includegraphics[width=\linewidth]{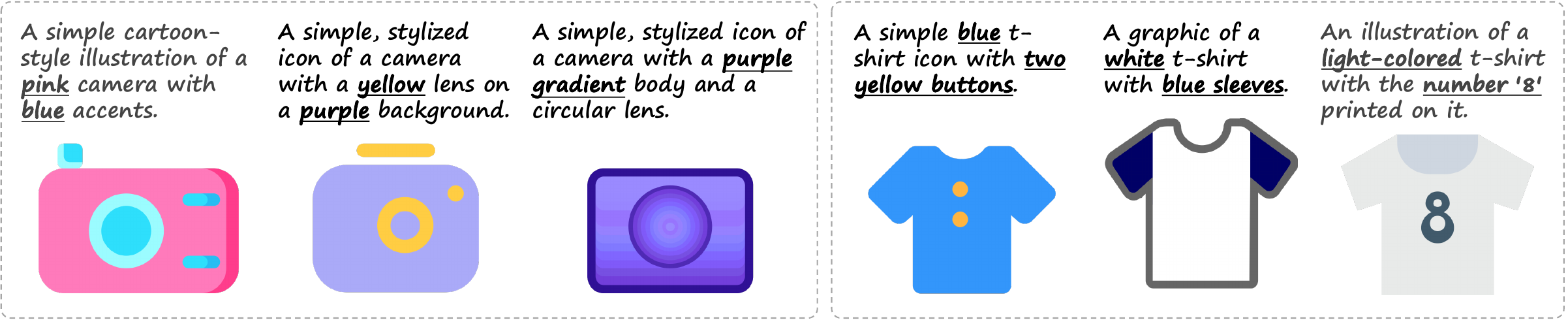}
    \caption{Instruction-following examples. Our model accurately applies specific attributes (\eg, colors and numbers) to their corresponding objects based on complex prompts.}
    \label{fig:instruction_follow}
\end{figure}

\begin{figure}[h]
    \centering
    \includegraphics[width=\linewidth]{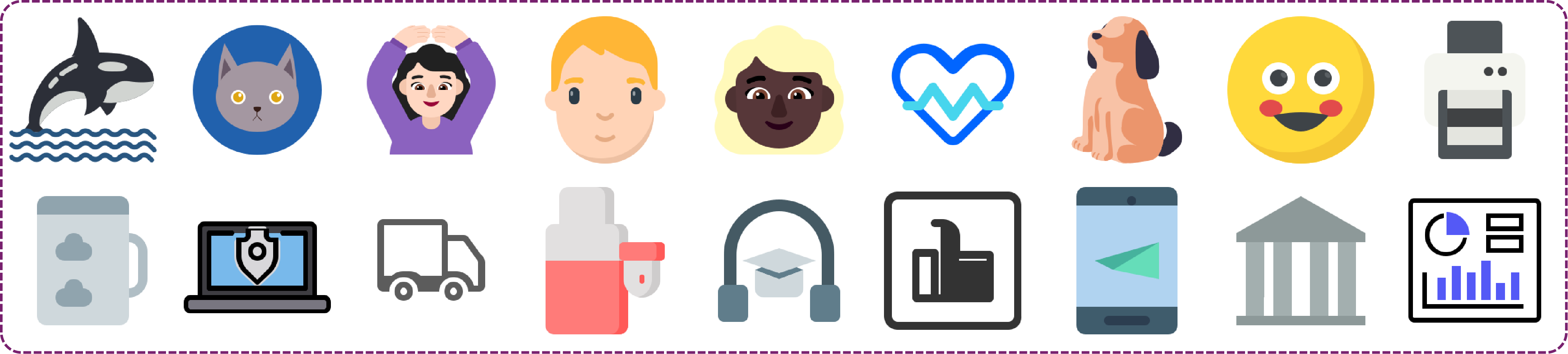}
    \caption{A gallery of diverse vector graphics generated by our Render-in-the-Loop model. Our approach consistently produces aesthetically pleasing and semantically coherent SVGs across a wide variety of subjects, maintaining topologically clean path structures.}
    \label{fig:gallery}
\end{figure}

We provide additional qualitative results that complement the main paper. \cref{fig:instruction_follow} shows fine-grained instruction-following examples, where our model accurately binds specific attributes (\eg, colors and counts) to their corresponding object parts. \cref{fig:gallery} further presents a gallery of diverse vector graphics generated by our Render-in-the-Loop model, illustrating its versatility across a wide variety of subjects while maintaining topologically clean path structures.

\section{Implementation Details}
\label{sec:supp_impl_details} 

\paragraph{Render and Verify.}
The Render-and-Verify (RaV) strategy is implemented as a lightweight inference-time verification module that filters degenerate generation steps. While the main paper describes the conceptual framework, here we provide additional implementation details for reproducibility.

At inference step $t$, the model first produces a candidate SVG code fragment $\hat{C}_t$. Before accepting this fragment, we render a hypothetical future canvas state
\[
\hat{I}_t = R(C_{1:t-1} \oplus \hat{C}_t),
\]
where $R(\cdot)$ denotes the SVG rasterization function and $C_{1:t-1}$ represents all previously accepted primitives.

Two heuristic checks are then applied.

\textbf{Visual Difference Check.}
We compute the pixel-wise difference between the rendered candidate canvas $\hat{I}_t$ and the previous canvas state $I_{t-1}$:

\[
\Delta(I_{t-1}, \hat{I}_t) =
\frac{1}{HW}
\sum_{i=1}^{H}\sum_{j=1}^{W}
|I_{t-1}(i,j) - \hat{I}_t(i,j)|.
\]

If $\Delta(I_{t-1}, \hat{I}_t) < \epsilon$, the candidate fragment is considered to provide negligible visual contribution and is rejected. In practice, this mechanism effectively filters redundant operations such as repeatedly drawing the same primitive, generating elements that are fully occluded, or placing shapes outside the visible canvas.

\textbf{Repetition Check.}
To prevent autoregressive repetition loops, we compute the string similarity between the candidate fragment $\hat{C}_t$ and the previously accepted fragment $C_{t-1}$. The similarity is measured using a normalized sequence matching score. If the similarity exceeds a threshold $\tau_{\text{sim}}$, the candidate fragment is rejected to avoid structurally redundant SVG primitives.

\textbf{Adaptive Resampling.}
If the candidate $\hat{C}_t$ fails either verification check, the fragment is discarded and the model resamples a new candidate. During resampling, the sampling hyperparameters (e.g., temperature and nucleus sampling range) are slightly increased to encourage the model to escape local repetitive distributions and explore alternative drawing actions. This process is repeated until a verified fragment is obtained.

\textbf{Termination Criterion.}
If the model fails to produce a valid fragment after a predefined maximum number of retries ($K_{\max}$), the system assumes that the drawing is already complete and forces the generation of the $\langle \text{END} \rangle$ token to terminate the drawing process.

In our implementation, all intermediate canvases are rasterized at a resolution of $224\times224$. The SVG rasterization is implemented using the CairoSVG renderer, which faithfully follows the W3C SVG specification~\cite{svg_w3c_1999}. During inference, we set the visual difference threshold to $\epsilon=0.001$ and the repetition similarity threshold to $\tau_{\text{sim}}=0.98$.

\begin{figure}[t]
\centering
\begin{tcolorbox}[
    colback=blue!5!white,
    colframe=blue!75!black,
    title=Text-to-SVG System Prompt,
    fonttitle=\bfseries,
    left=5pt, right=5pt, top=5pt, bottom=5pt,
    sharp corners,
    boxrule=0.5pt
]
\small
You are an expert SVG artist and coder. Your task is to generate Scalable Vector Graphics (SVG) code incrementally based on a visual feedback loop. \\
\\
\textbf{\#\# Workflow Protocol} \\
1. \textbf{Initialization}: The user will provide a text description. \\
2. \textbf{Incremental Drawing}: Do NOT output the entire SVG at once. Output small, logical fragments (e.g., one shape or path) in each turn wrapped in \texttt{```svg} code blocks. \\
3. \textbf{Visual Self-Correction}: \\
\hspace{1em} - After your first turn, the user will \textbf{STOP} providing text instructions. \\
\hspace{1em} - The user will \textbf{ONLY} input a rasterized \texttt{image} of the current canvas state. \\
\hspace{1em} - IMPORTANT: Each \texttt{image} is the PNG render of the SVG canvas produced by cumulatively applying all SVG fragments you have output so far. Treat it as the current accumulated drawing state. \\
\hspace{1em} - You must act as the ``eye'' and the ``hand'': Look at the image, and output the \textit{next} SVG code fragment to complete the drawing. \\
4. \textbf{Termination}: \\
\hspace{1em} - When you see the image and determine the drawing is fully complete and matches the goal, you must output a special termination signal inside an svg block: \\
\hspace{1em} \texttt{```svg} \\
\hspace{1em} \texttt{<END>} \\
\hspace{1em} \texttt{'''} \\
\\
\textbf{\#\# Technical Constraints} \\
- Canvas Size: 224x224. \\
- Coordinate System: Maintain strict spatial awareness within the 0-224 range. \\
- Style: Concise, geometric, and aesthetically pleasing vector art.
\end{tcolorbox}
\vspace{-1em}
\caption{System Prompt for Text-to-SVG Task.}
\label{prompt:t2s}
\vspace{-1em}
\end{figure}

\begin{figure}[!t]
\centering
\begin{tcolorbox}[
    colback=blue!5!white,
    colframe=blue!75!black,
    title=Image-to-SVG System Prompt,
    fonttitle=\bfseries,
    left=5pt, right=5pt, top=5pt, bottom=5pt,
    sharp corners,
    boxrule=0.5pt
]
\small
You are an expert SVG artist and visual reverse-engineering model. Your task is to reconstruct Scalable Vector Graphics (SVG) code incrementally based on a visual feedback loop. \\
\\
\textbf{\#\# Task Definition} \\
You are given an image that represents the final target appearance of an SVG drawing. Your goal is to reproduce this image using SVG code, generated step by step. \\
\\
\textbf{\#\# Workflow Protocol} \\
1. \textbf{Initialization} \\
\hspace{1em} - The user's first input will be an image. \\
\hspace{1em} - Observe the image and infer the intended visual result. \\
2. \textbf{Incremental Drawing} \\
\hspace{1em} - Do NOT output the entire SVG at once. \\
\hspace{1em} - Output small, logical SVG fragments wrapped in \texttt{```svg} code blocks. \\
\hspace{1em} - All SVG fragments are cumulatively applied to the same canvas. \\
3. \textbf{Visual Feedback Loop} \\
\hspace{1em} - After your first SVG output, the user will stop providing instructions. \\
\hspace{1em} - The user will ONLY provide images as input. \\
\hspace{1em} - Each image represents the current canvas state rendered from all SVG fragments so far. \\
\hspace{1em} - Compare the image with your intended result and output the next SVG fragment needed. \\
4. \textbf{Termination} \\
\hspace{1em} - When the drawing is complete and matches the target image, output: \texttt{<END>} \\
\\
\textbf{\#\# Technical Constraints} \\
- Canvas size: 224 $\times$ 224. \\
- Coordinate system: 0--224. \\
- Style: Concise, geometric, and visually faithful.
\end{tcolorbox}
\vspace{-1em}
\caption{System Prompt for Image-to-SVG Task.}
\label{prompt:i2s}
\vspace{-1em}
\end{figure}

\begin{figure}[t]
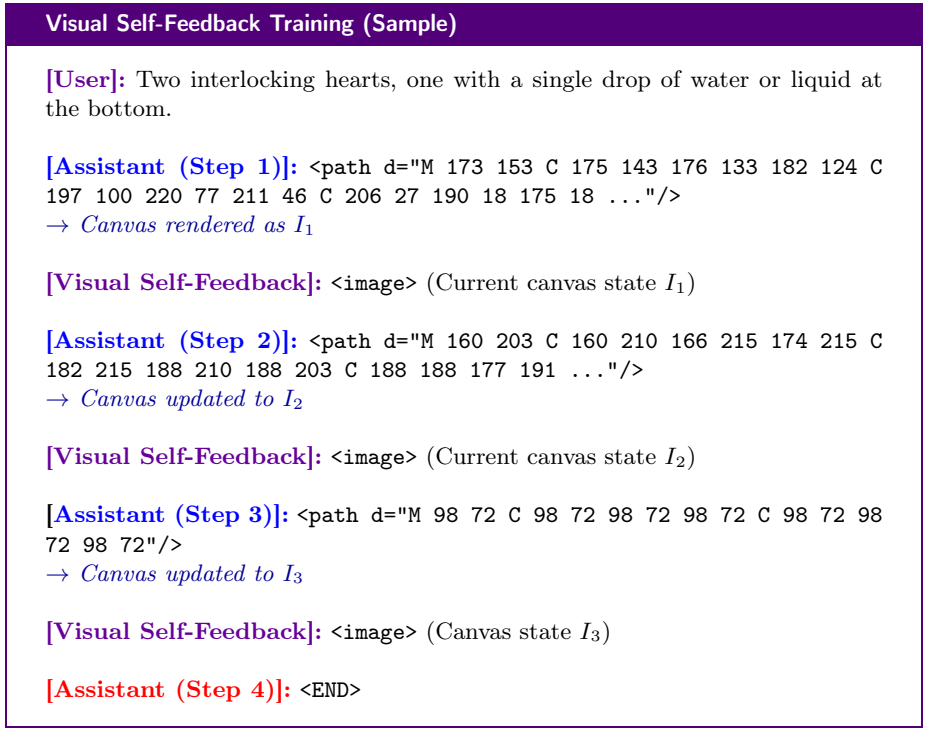

\centering
\begin{tcolorbox}[
    colback=white,
    colframe=MainPurple!80!black,
    title=\textbf{\sffamily Visual Self-Feedback Training (Sample)},
    fonttitle=\bfseries\small,
    boxrule=0.8pt,
    sharp corners
]
\small
\textbf{\color{MainPurple} [User]:} Two interlocking hearts, one with a single drop of water or liquid at the bottom. \\

\textbf{\color{blue} [Assistant (Step 1)]:} \texttt{<path d="M 173 153 C 175 143 176 133 182 124 C 197 100 220 77 211 46 C 206 27 190 18 175 18 ..."/>} \\
\textit{\color{blue!60!black} \footnotesize $\rightarrow$ Canvas rendered as $I_1$} \\

\textbf{\color{MainPurple} [Visual Self-Feedback]:} \texttt{<image>} (Current canvas state $I_1$) \\

\textbf{\color{blue} [Assistant (Step 2)]:} \texttt{<path d="M 160 203 C 160 210 166 215 174 215 C 182 215 188 210 188 203 C 188 188 177 191 ..."/>} \\
\textit{\color{blue!60!black} \footnotesize $\rightarrow$ Canvas updated to $I_2$} \\

\textbf{\color{MainPurple} [Visual Self-Feedback]:} \texttt{<image>} (Current canvas state $I_2$) \\

\textbf{[\color{blue} Assistant (Step 3)]:} \texttt{<path d="M 98 72 C 98 72 98 72 98 72 C 98 72 98 72 98 72"/>} \\
\textit{\color{blue!60!black} \footnotesize $\rightarrow$ Canvas updated to $I_3$} \\

\textbf{\color{MainPurple} [Visual Self-Feedback]:} \texttt{<image>} (Canvas state $I_3$) \\

\textbf{\color{red} [Assistant (Step 4)]:} \texttt{<END>} 
\end{tcolorbox}
\vspace{-1.5em}
\caption{A sample from our fine-grained decomposition dataset. The Assistant generates SVG segments incrementally, interleaved with rendered visual context.}
\label{fig:data_sample}
\vspace{-1em}
\end{figure}
\paragraph{System Prompt.}
We present the system prompts used for Text-to-SVG and Image-to-SVG tasks in \cref{prompt:t2s} and \cref{prompt:i2s}, respectively. These prompts guide the MLLM to follow the iterative drawing protocol and leverage visual feedback.
\paragraph{Dataset Construction Examples.}
To illustrate the Visual Self-Feedback (VSF) training format, we provide a representative training sample in \cref{fig:data_sample}. The sequence demonstrates how the model learns to map instructions to incremental SVG paths and consistently conditions its next output on the rendered visual feedback.

\section{More Details of the Baselines}
\label{sec:morebaselines}
In this section, we provide additional details of the baseline methods used in our experiments. These methods span three categories: optimization-based approaches, autoregressive LLM/VLM-based models, and reinforcement learning based methods.

\paragraph{Optimization-based Methods.}

\textbf{VectorFusion}~\cite{vectorfusion_jain_2023} generates SVG graphics from text prompts by distilling knowledge from pretrained diffusion models into vector representations. It optimizes vector parameters using score distillation sampling (SDS) together with a differentiable vector graphics renderer. While the method can produce diverse vector graphics without large SVG training datasets, the optimization process is computationally expensive and often requires long inference time.

\textbf{SVGDreamer}~\cite{svgdreamer_xing_2024} proposes a semantic-driven image vectorization (SIVE) framework for text-to-SVG generation. It separates foreground objects from background regions and optimizes vector primitives using a vectorized particle-based score distillation strategy. Although this approach improves editability and semantic structure, the iterative optimization pipeline remains computationally intensive.

\textbf{LIVE}~\cite{live_ma_2022} (Layer-wise Image Vectorization) reconstructs SVG graphics by progressively adding and optimizing vector paths to match a raster target image. The method employs a differentiable renderer to optimize path parameters and introduces component-wise path initialization to improve topology extraction.

\textbf{DiffVG}~\cite{diffvg_li_2020} is a differentiable vector graphics rasterization framework that enables gradient-based optimization over vector primitives. By computing gradients through the rasterization pipeline, DiffVG directly optimizes vector parameters to match raster images. Despite its effectiveness in image vectorization tasks, the forward–backward rasterization process introduces significant computational overhead.

\paragraph{Autoregressive LLM/VLM-based Methods.}

\textbf{StarVector}~\cite{starvector_rodriguez_2025} generates SVG code directly from images using a multimodal transformer architecture. It integrates an image encoder with a language model to autoregressively predict SVG primitives. While the model demonstrates strong performance on icon-level vectorization tasks, its limited context length restricts the generation of highly complex SVG structures.

\textbf{IconShop}~\cite{iconshop_wu_2023} adopts a transformer-based autoregressive model to generate SVG path sequences. By tokenizing SVG commands and coordinates, the model learns to synthesize vector graphics from textual descriptions. IconShop achieves competitive performance for icon synthesis but is mainly designed for relatively simple and monochrome SVG structures.

\textbf{GPT-5}~\cite{GPT-5} represents a large-scale multimodal language model capable of generating SVG code through instruction following. It demonstrates strong general reasoning and visual understanding capabilities but is not specifically optimized for structured vector graphics generation.

\textbf{OmniSVG}~\cite{omnisvg_yang_2025} is a unified multimodal SVG generation framework built upon a vision-language model backbone. By parameterizing SVG commands and coordinates into discrete tokens, OmniSVG enables autoregressive SVG generation conditioned on textual or visual inputs.

\textbf{InternSVG}~\cite{wang2025internsvg} is a recent large-scale SVG generation model trained on the SAgoge dataset containing approximately 16M SVG samples. It leverages large multimodal transformers to generate complex vector graphics and demonstrates strong scalability with large training corpora.

\paragraph{Reinforcement Learning-based Methods.}

\textbf{SVGen}~\cite{svgen_wang_2025} introduces reinforcement learning for SVG generation by optimizing the generation policy using the Group Relative Policy Optimization (GRPO)~\cite{guo2025deepseek} algorithm. The method aims to improve the alignment between generated SVG outputs and evaluation metrics through policy optimization. As one of the first RL-based approaches for SVG generation, it provides an important baseline for evaluating reinforcement learning strategies in vector graphics synthesis.

Unless otherwise specified, we use the official implementations of these baselines and follow the hyperparameters recommended in their original papers.

\section{Comparison with LIVE and DiffVG}
\label{sec:showcase_baselines}
While optimization-based vectorization methods such as LIVE~\cite{live_ma_2022} and DiffVG~\cite{diffvg_li_2020} can achieve high visual fidelity through gradient descent, their resulting SVG representations often lack human-interpretable structure and logical layering. 

\begin{figure}[t]
\centering
\includegraphics[width=0.8\textwidth]{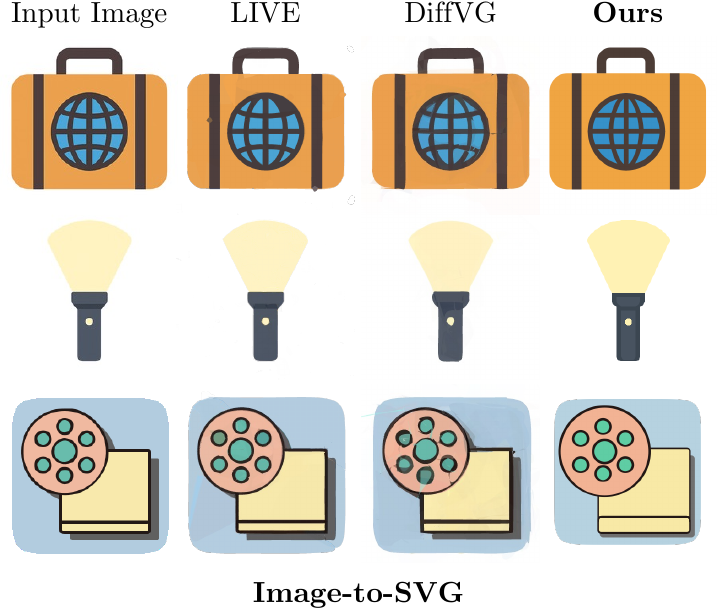}
\caption{A zoomed-in visualization of the SVG paths generated by optimization-based baselines (LIVE and DiffVG). To minimize the pixel-wise rendering loss, these methods tend to produce chaotic, densely overlapping B\'ezier curves and redundant shape layers. This results in an uneditable and semantically structureless vector representation, which starkly contrasts with the clean, logically constructed outputs produced by our approach.}
\label{fig:diss_baselines}
\vspace{-1em}
\end{figure}

As demonstrated in \cref{fig:diss_baselines}, these methods typically generate a massive number of redundant paths and messy overlapping layers to merely fit the target image pixels. This ``stroke-stacking'' behavior leads to bloated SVG files that are difficult for human designers to edit, animate, or reuse. In contrast, our proposed \textbf{Render-in-the-Loop} framework leverages the semantic prior of Multimodal Large Language Models (MLLMs) to generate clean, geometrically logical, and highly editable vector graphics that align with human drawing habits.

\section{MMSVGBench}
\label{sec:mmsvgbench}
We evaluate our method on MMSVGBench~\cite{omnisvg_yang_2025}, a benchmark proposed by OmniSVG~\cite{omnisvg_yang_2025} for evaluating multimodal SVG generation models.
The benchmark contains 600 synthetic samples covering two tasks: text-to-SVG and image-to-SVG.
Each task contains 300 evaluation samples, further divided into icon-level and illustration-level complexity.

To ensure fair evaluation and avoid training data leakage, all prompts and images in MMSVGBench are generated from scratch using GPT~\cite{gpt4_achiam_2023} models.
Therefore, the benchmark samples are guaranteed to be unseen during model training.

Each benchmark sample contains the input modality (text or image), task type, semantic category (icon or illustration), and metadata such as the source URL.
Following the protocol of OmniSVG, we report FID, CLIP score, Aesthetic score and HPS for text-to-SVG tasks, and DINO similarity, SSIM, LPIPS, and MSE for image-to-SVG tasks.

\section{Related Works}

\paragraph{Optimization and Autoregressive SVG Generation.}
Traditional vector graphics generation~\cite{clipdraw_frans_2022,wordasimage_iluz_2023,nivel_thamizharasan_2024,t2vecneualpath_zhang_2024,sketchagent_vinker_2025,samvg_zhu_2024,supersvg_hu_2024,sglive_zhou_2024,optimizeandreduce_hirschorn_2024,adavec_zhao_2025,layered_wang_2025,layerpeeler_wu_2025,mosketch_liu_2025,livesketch_gal_2024,wang2026reliable} is often formulated as an optimization problem. Enabled by differentiable rasterizers~\cite{diffvg_li_2020}, early approaches~\cite{live_ma_2022, clipasso_vinker_2022,clipascene_vinker_2023} focus on image-to-SVG reconstruction. Subsequent works combine text-to-image diffusion models~\cite{ldm_rombach_2022, ddpm_ho_2020} with Score Distillation Sampling (SDS)~\cite{dreamfusion_poole_2023} to achieve text-to-SVG generation~\cite{vectorfusion_jain_2023, svgdreamer_xing_2024, diffsketcher_xing_2023}. While visually appealing, these optimization-based methods~\cite{vectorfusion_jain_2023,groupsketch_liang_2025,viewcraft3d_wang_2025,svgdreamer++_xing_2025,vectorpainter_hu_2025} suffer from excessive inference latency and often produce chaotic paths that sacrifice topological editability. 
To bypass optimization bottlenecks, recent works treat SVG generation as a direct sequence modeling task. Leveraging large language models (LLMs)~\cite{gpt4_achiam_2023, qwen3vl_bai_2025}, these methods synthesize SVGs by autoregressively predicting XML tags and path coordinates~\cite{iconshop_wu_2023, starvector_rodriguez_2025, omnisvg_yang_2025, wang2025internsvg, llm4svg_xing_2025,zhang2025duetsvg,deepsvg_carlier_2020,sketchrnn_david_2018}. However, these autoregressive models typically operate in an open-loop manner, blindly decoding abstract coordinates without observing the intermediate rendering results, which frequently leads to geometric hallucinations and structural inaccuracies.

\paragraph{Feedback-Driven SVG Generation.}
To mitigate the limitations of open-loop autoregressive generation, feedback mechanisms have been introduced. In broader text and code generation tasks, reinforcement learning (RL) techniques~\cite{schulman2017proximal, guo2025deepseek} leverage outcome-based reward models~\cite{imagereward_xu_2023} to align outputs with human preferences or logical constraints. In the specific context of vector graphics, recent methods such as SVGen~\cite{svgen_wang_2025}, Reason-SVG~\cite{reasonsvg_xing_2025}, and rendering-aware RL approaches~\cite{rlrf_rodriguez_2025} explore the use of verifiable reinforcement learning. These methods typically employ rendering validity and visual-semantic alignment (\eg, CLIP~\cite{clip_radford_2021} scores) as scalar rewards to optimize the generated SVG codes. While RL introduces delayed guidance, mapping complex visual structures into unidimensional scalar rewards inherently compresses away critical spatial information. 
In this paper, we explore a more direct dense feedback mechanism. Instead of relying on abstract scalar rewards, our proposed Render-in-the-Loop paradigm relies on constructing Visual Self-Feedback (VSF) training sequences. By continuously rendering and injecting intermediate canvas states back into the vision pathway of Multimodal Large Language Models (MLLMs), we provide precise step-by-step visual guidance, achieving highly competitive vector synthesis while avoiding the spatial information loss typical of reward-based scalar feedback. We emphasize that VSF is not a replacement for RL but a complementary mechanism: VSF reshapes the generation \emph{state} by exposing intermediate visual context, whereas RL optimizes the \emph{policy} over such trajectories. The two are naturally compatible, and we leave their combination to future work.

\paragraph{Broader Context in Multimodal and Generative Learning.}
Beyond vector graphics, our work connects to the broader progress in multimodal perception and structured visual generation. On the generation side, recent methods tackle layout and advertising-image synthesis with human feedback~\cite{lu2025uni,lu2026one}, as well as 3D scene reconstruction and physically grounded world synthesis~\cite{pan2024harmonicnerf,lu2026worldcoder}, all of which, like ours, benefit from injecting structured or visual feedback into the generation process. On the perception side, motion- and geometry-centric models for optical flow, scene flow, and point-cloud registration~\cite{liu2026arflow,liu2024difflow3d,liu2023regformer}, together with flow-matching and reinforcement-learning-based trajectory optimization~\cite{tian2026curvatureadaptiveconsistencyflowmatching}, highlight the value of progressive, iterative refinement that resonates with our step-wise drawing formulation. Finally, advances in efficient and multimodal foundation models~\cite{guo2025quantized,tao2025autopcr}, agent-based reasoning and data synthesis~\cite{liu2026memory,yu2026mathagentadversarialevolutionconstraint}, efficient learning strategies such as knowledge distillation, sampler optimization, neuroevolution, and preference-driven code generation~\cite{li2025frequency,yao2024swift,li2026evolving,li2025preference}, and human-centered multimodal applications spanning virtual-reality therapy and health informatics~\cite{zhou2025shadow,zhou2025adhera} provide complementary tools and insights that systems like ours can build upon. Our Render-in-the-Loop paradigm shares the spirit of these works in exploiting structured feedback and strong visual priors, while focusing specifically on closing the rendering loop for SVG generation.

\label{sec:supp_related}

\end{document}